\documentclass[runningheads]{llncs}
\usepackage[T1]{fontenc}
\usepackage{graphicx}
\usepackage{multirow}
\usepackage[table]{xcolor}
\usepackage{booktabs}
\usepackage{makecell}
\usepackage{placeins}
\usepackage{orcidlink}

\usepackage{amsmath,amssymb}
\usepackage{algorithm}
\usepackage{algpseudocode}
\usepackage[misc]{ifsym} 

\newcommand{\corr}{(\Letter)}
\usepackage{mwe}

\begin{document}

\title{Neural Network Compression by Approximate Differential Equivalence}

\titlerunning{Neural Network Compression by Approximate Differential Equivalence}

\author{Ravi Dhiman\inst{1,2}\orcidlink{0009-0005-6595-2621}\corr \and
Andrea Passarella \inst{2}\orcidlink{0000-0002-1694-612X} \and
Mirco Tribastone\inst{1}\orcidlink{0000-0002-6018-5989} \and
Lorenzo Valerio\inst{2}\orcidlink{0000-0001-5574-7847}}

\authorrunning{R. Dhiman et al.}

\institute{}
\institute{IMT School for Advanced Studies Lucca 55100, Italy\\\email{\{ravi.dhiman, mirco.tribastone\}@imtlucca.it}
\and
IIT CNR, Pisa 56124, Italy\\ \email{\{andrea.passarella, lorenzo.valerio\}@iit.cnr.it}}

\maketitle              
\begin{abstract}
Neural network compression is commonly achieved by pruning parameters based on local importance scores, e.g., magnitude-based pruning. We propose a complementary approach that compresses models by aggregating neurons with similar functional behavior rather than removing weights independently. Our method encodes a trained network as a polynomial ODE system and applies a lumping method called Approximate Forward Differential Equivalence to identify neurons with approximately matching induced dynamics. A single tolerance parameter, $\varepsilon$, controls the compression level and induces a smooth trade-off between model size and predictive accuracy. We evaluate the method on synthetic datasets derived from nonlinear dynamical systems with known ground-truth behavior and on public regression benchmarks. Across both settings, the proposed approach achieves substantial parameter reduction while preserving accuracy, and consistently compares favorably with magnitude-based pruning and Wanda at similar compression levels. These results suggest that differential equivalence-based aggregation is a principled and effective alternative to conventional weight-centric pruning.

\end{abstract}
\section{Introduction}
Performing AI tasks, particularly training complex AI systems (such as deep neural networks, or DNNs) on resource-constrained devices, is a cutting-edge research area that presents numerous stimulating methodological and technological challenges.
DNNs have played a key role in fields such as natural language processing~\cite{devlin2019bert}, computer vision~\cite{dosovitskiy2021image,simonyan2015very}, and cross-modal applications~\cite{lin2024vila,liu2024sora}. These models have shown tremendous
performance, primarily due to their complex architectures and large number of parameters~\cite{cheng2024ieee,devlin2019bert,he2016deep}. 
However, this success comes at a cost. Neural networks require significant energy consumption~\cite{dong2017more,han2016deepcompression,you2019gate} due to their computational and memory-intensive nature. Due to these requirements, these models are complex to deploy on devices that are resource-constrained, such as mobile phones, edge devices, and embedded systems \cite{luo2017thinet}. Furthermore, various challenges, such as slower inference times and higher latency, arise in real-time applications like remote sensing, autonomous driving, and emergency response systems.  

Oftentimes, trained NNs are largely over-parameterized, i.e., the same accuracy can be achieved with smaller networks with fewer parameters. A common method to tackle this problem is to simplify neural networks by pruning unnecessary parameters or groups of them. The design space of pruning algorithms is rich, spanning \emph{what} to prune (schemes over weights, channels, blocks, or layers)\cite{he2017channel,luo2017thinet}, \emph{how much} to prune (layer-wise sparsity allocation)~\cite{lee2020layer,sanh2020movement}, \emph{which} parameters to select\cite{he2019filter,orseau2020logarithmic}, and \emph{how} to train (prune-then-finetune vs.\ train-time sparsification)\cite{wang2020neural}. Selection criteria range from classical magnitude-based rules~\cite{he2019filter,lee2020layer} to gradient- and sensitivity-driven scores~\cite{liu2021group}.
Depending on the structure removed, pruning methods are typically categorized as \emph{unstructured pruning}~\cite{frankle2019lottery,sun2024pruning,tanaka2020synaptic}, \emph{structured pruning}~\cite{ma2023llmpruner,wang2023trainability}, and \emph{semi-structured pruning}~\cite{frantar2023sparsegpt,ma2020real,meng2020filterinfilter}. In \emph{dependency-based pruning} models, the parameters are not independent but rather have complex interrelationships~\cite {liu2021group,you2019gate,zhang2021aligned}. In such models, removing or modifying a weight or neuron can affect many downstream computations because of shared activations, skip connections, or dense feature reuse, making this an ongoing challenge in modern architectures. To preserve accuracy after pruning, all regimes require careful fine-tuning, stability controls, and sensitivity-aware scheduling. Inspired by the theory of Markov chains, some recent compression methods consider lumpability~\cite{kemeny1976finite}, where redundant components are grouped without sacrificing performance. Neural lumping follows state lumping in Markov chains and reduces topology while preserving information flow~\cite{ressi2022lumping,ressi2024formal}.

 In this work, we investigate a complementary approach to model simplification by means of lumping, which performs parameter \emph{aggregation} rather than removal. In contrast to weight-centric pruning rules, lumpability-based approaches take a step back and ask a different question: which parts of the network are effectively doing the same thing? These methods follow a more structured approach, first identifying neurons with nearly identical behavior and then merging them, thereby reducing the model size. In this way, compression depends less on hand-crafted heuristics and repeated fine-tuning, which is especially appealing when redundancy appears at the level of learned functionality rather than at the level of individual parameters.

Concretely, we introduce a compression approach based on an approximate forward differential equivalence ($\varepsilon$-FDE). The method aggregates neurons whose induced dynamics are approximately the same.
This shifts the focus from the importance of local parameters to functional similarity within the network. The tolerance parameter $\varepsilon$ controls how aggressively neurons are merged, offering a clear and continuous trade-off between compression and accuracy. We evaluate the developed method using synthetic datasets derived from nonlinear dynamical systems (i.e., reaction networks, electrical networks, polymerization dynamics, and glycolytic oscillators) with known ground-truth dynamics, which allows us to study aggregation effects in a controlled setting and on public benchmarks including four regression datasets (Abalone, Metro Interstate Traffic, Individual Household Electric Power Consumption, and Protein).
Experiments on both synthetic dynamical benchmarks and real-world regression datasets show that aggregating functionally similar components can significantly reduce model size while maintaining accuracy, and do so more stably than standard pruning baselines such as magnitude-based pruning and Wanda. Specifically, our method consistently achieves a significant reduction in parameters (often exceeding 60\%) while keeping the test MSE close to that of the original model. At comparable compression levels, MBP and Wanda typically achieve MSEs of orders of magnitude higher.

 The paper is structured as follows. In Section~2, we provide the necessary background to understand the adopted methodology. Section~3 formally describes the methodology. In Section~4, we present the results and discussion. We conclude and highlight some future work directions in Section~5.
 
\section{Background}
The central idea of our approach is to simplifiy a trained neural network by \emph{aggregating similar behaved neurons}, rather than removing individual parameters. To find similarities, we encode the network as a system of ordinary differential equations (ODEs) and look for (approximate) symmetries in the resulting dynamics. Symmetries allow neurons to be grouped into the same component while preserving the evolution of their joint activity up to a predetermined tolearance parameter $\varepsilon$.

We begin with an ODE example to understand the basis of the adopted methodology.
Consider the following system of ordinary differential equations:
\begin{align}
\frac{d x_1}{d t} &= -5.00 x_1 + x_2 + x_3, \\
\frac{d x_2}{d t} &= 2.99 x_1 - x_2, \\
\frac{d x_3}{d t} &= 2.01 x_1 - x_3.
\end{align}
Although $x_2$ and $x_3$ are distinct variables, their dynamics are
structurally similar. Indeed, the sum $x_2 + x_3$ evolves autonomously:
\[
\frac{d}{dt}(x_2 + x_3) = 5.00 x_1 - (x_2 + x_3).
\]
This observation suggests that $\{x_2, x_3\}$ can be treated as a single
group of variables, referred to as an \emph{aggregated block-sum} variable (or simply a \emph{block-sum} variable)\cite{cardelli2016symbolic}, and tracked through their sum. Accordingly, the block-sum $x_2+x_3$ defines a block-sum variable $x_{23}$, which replaces the individual states $x_2$ and $x_3$ in the reduced description.

The formalization of this idea is called \emph{Forward differential equivalence} (FDE) \cite{cardelli2023formal,cardelli2016symbolic}. Specifically, a partition of ODE variables is an FDE if, for each block\footnote{In this paper we use the terms \emph{block} to indicate a group of variables.}, the sum of the variables in
that block satisfies a closed differential equation. Here, for the given system of ODEs, partition of variables $\{\{x_1\}, \{x_2,x_3\}\}$ is a FDE, since the dynamics of the block $\{x_2,x_3\}$ can be expressed in terms of the block-sum variable $x_{23} = x_2 + x_3$, and the resulting
\emph{quotient ODE system} is 
\begin{align}
\frac{d x_1}{d t} &= -5.00 x_1 + x_{23}, \\
\frac{d x_{23}}{d t} &= 5.00 x_1 - x_{23}.
\end{align}
This quotient system evolves these block-sums directly, and its solutions exactly coincide with the sums of the original variables at all times, provided the initial sums match. This aggregation entails a loss of resolution, since we cannot observe the individual variables $x_2$ and $ x_3$ separately; only their combined contribution, $ x _ {23} $, is retained. This is an inherent consequence of quotienting under FDE, which preserves block-sum variables rather than individual variables. In our setting, this loss of detail is not critical, since the dynamics of interest depend only on the combined effect of variables within each block. As a result, the reduced system preserves the relevant behavior while providing a simpler representation.

However, it must be noted that exact symmetries are fragile, because even small perturbations of coefficients might break them.  
For instance, modifying the system above to
\begin{align}
\frac{d x_2}{d t} &= 2.99 x_1 - 1.05 x_2, \\
\frac{d x_3}{d t} &= 2.01 x_1 - 1.00 x_3.
\end{align}
breaks exact closure as the variables $x_2$ and $x_3$ are no longer exactly interchangeable due to slightly differing interaction coefficients (1.05 vs. 1.00). Thus,
exact FDE no longer holds. 
However, if these differences remain below a chosen
tolerance, we can still group these variables as approximately equivalent. In this way, the system can be treated as an \emph{approximate FDE} \cite{cardelli2023formal}. 

The approximate FDE formally relaxes the notion of symmetry by
allowing bounded discrepancies in coefficients. The tolerance parameter
$\varepsilon$ quantifies the admissible mismatch: if the induced error
on block-sum derivatives is bounded by $\varepsilon$, the block is
accepted. 
In the modified case, if we choose
$\varepsilon \geq 0.05$, the grouping \{\{$x_1$\}, \{$x_2, x_3$\}\} is still considered approximately equivalent. We define the \emph{approximate quotient system} by summing the approximately
equivalent variables, setting $x_{23} = x_2 + x_3$:
\begin{align}
\frac{d x_1}{d t} &= -5.00 x_1 + x_{23}, \\
\frac{d x_{23}}{d t} &\approx 5.00 x_1 - x_{23}.
\end{align}

These notions generalize to Polynomial Initial Value Problems (PIVPs) \cite{cardelli2023formal,cardelli2016symbolic},
i.e., systems of ODEs of the form
\begin{align}
\frac{d x_i}{d t} = q_i(x), \quad i = 1,\dots,n,
\end{align}
where each $q_i$ is a multivariate polynomial. PIVPs form a well-studied class for which both exact and approximate differential equivalences can be computed effectively via partition refinement.

We now formally define the approximate FDE for the polynomial ODE system obtained from the neural network encoding.

\begin{definition}[Approximate FDE]
Let $\mathcal{S}$ be the state space of the polynomial ODE system obtained
from the encoding of a trained neural network
$\mathcal{N}$. A partition $P=\{P_1,\dots,P_k\}$ of $\mathcal{S}$ is an
approximate forward differential equivalence (approximate FDE)
partition if, for any block $P_m\in P$ and any pair of variables
$x_i,x_j\in P_m$, the induced discrepancy in block-sum derivatives is
uniformly bounded by $\varepsilon$, i.e.,
\[
\sum_{\alpha} \bigl| c(\varphi^{H}_{i,j}, x^\alpha) \bigr| \le \varepsilon
\qquad \text{for all } H \in P,
\]
where $\varphi^{H}_{i,j}$ denotes the polynomial capturing the difference
in the contribution of variables $x_i$ and $x_j$ to the derivative of the
block-sum associated with block $H$, under redistributions that preserve
the total value within $H$.
The notation $c(\varphi^{H}_{i,j}, x^\alpha)$ refers to the coefficient of
the monomial $x^\alpha$ in this discrepancy polynomial.
The partition $P$ is obtained as the transitive closure of this relation
on $\mathcal{S}$~\cite{cardelli2023formal}.
\end{definition}

In the discussed ODE example, consider the block $H=\{x_2,x_3\}$ and the partition $P = \{\{x_1\},\{x_2,x_3\}\}$, then the discrepancy polynomial $\varphi^{H}_{2,3}$ corresponds to the difference between the contributions of $x_2$ and $x_3$ to the derivative of the
block-sum $x_{23}=x_2+x_3$, as induced by the right-hand sides of
$\frac{d x_2}{d t}$ and $\frac{d x_3}{d t}$.

\paragraph{Roadmap.}
Table~\ref{tab:roadmap} summarizes the adopted methodology's roadmap. Each step is detailed formally in Section 3.

\begin{table}[t]
\centering
\caption{Conceptual roadmap from a trained neural network to its aggregated form.}
\label{tab:roadmap}
\begin{tabular}{llp{6.8cm}}
\toprule
Step & Object & Role in the pipeline \\
\midrule
(1) & Neural network parameters
    & A trained neural network $\mathcal{N}$ with weights and biases. \\[0.6ex]

$\rightarrow$ & Polynomial ODE encoding $q$
    & The network is encoded exactly as a system of polynomial ODEs,
      capturing the layer-wise forward computation in a form amenable
      to differential equivalence analysis. \\[0.6ex]

$\rightarrow$ & $\varepsilon$-FDE partition $P$
    & Approximate FDE identifies groups of variables whose dynamics are approximately symmetric up to a tolerance $\varepsilon$. \\[0.6ex]

$\rightarrow$ & Quotient ODE $g$
    & The reference model $\hat{q}$ admits an exact quotient under $P$, yielding reduced dynamics evolving block-sum variables. \\[0.6ex]

$\rightarrow$ & Aggregated neural network $\mathcal{N}_{\text{agg}}$
    & An aggregated network is constructed whose polynomial ODE
      encoding coincides with the quotient dynamics. \\
\bottomrule
\end{tabular}
\end{table}

\section{Methodology}
\label{sec:methodology}
In this section, we present the proposed compression framework. First, we present the formal procedure, followed by a working example that we use to make it concrete and easily understood. 

To apply differential equivalence techniques, we first view a feedforward neural network as a (degenerate) dynamical system: the forward computation of each layer is represented as the right-hand side of an ODE.\footnote{This construction is not intended to model temporal evolution.
Rather, ODEs are used as a symbolic representation that enables the
application of algebraic symmetry-detection machinery to polynomial
maps.} Then, we compute an approximate FDE partition ($\varepsilon$-FDE) and aggregate the neurons based on the induced quotient dynamics. The evaluation of the resulting reduced model is presented in Section 4.

Although FDE enforces exact symmetries in polynomial
ODE systems, ensuring that sums of variables within each block satisfy closed quotient dynamics~\cite{cardelli2016symbolic}, trained neural networks rarely exhibit such exact algebraic structure. Neurons that play a similar functional role typically differ slightly in their learned
weights and biases, breaking the exact algebraic symmetry at the level of the induced dynamics. This mismatch motivates the use of approximate FDE which accommodates such small discrepancies~\cite{cardelli2023formal} and enables aggregation in realistic, trained
models.

We start with the formal definition of the polynomial ODE encoding of a trained network.
\begin{definition}[Polynomial ODE encoding of a trained network]
\label{def:ode-encoding}
Let $\mathcal{N}$ be a trained fully connected neural network with weight matrices $W^{(\ell)} \in \mathbb{R}^{n_\ell \times n_{\ell-1}}$ and bias vectors $b^{(\ell)} \in \mathbb{R}^{n_\ell}$ at layer $\ell$.
For each selected layer $\ell$, we introduce state variables
$x_i^{(\ell)}$ representing neuron outputs and define their dynamics by
\[
\frac{d x^{(\ell)}}{d t}
    = \sigma^{(\ell)}\!\left(W^{(\ell)} x^{(\ell-1)} + b^{(\ell)}\right),
\]
where $\sigma^{(\ell)}$ is a (polynomial) activation function applied
component-wise.
Collectively, the dynamics across all selected layers yield a
\emph{polynomial initial value problem} (PIVP), i.e., a system of ODEs
whose right-hand sides are multivariate polynomials.
The associated state space is
\[
\mathcal{S}
= \{\, x_i^{(\ell)} : \ell \in \mathcal{L},\; i = 1,\dots,n_\ell \,\}.
\]
\end{definition}

Note that the polynomial ODE encoding does not have a physical interpretation with respect to a neural network. Instead, this ODE system is used as a symbolic representation to identify
functionally similar neurons via differential equivalence and allows us to construct an equivalent reduced neural network in a principled way. We rewrite the resulting polynomial ODE encoding in a monomial normal form~\cite{cardelli2023formal}~\footnote{This transformation is semantics-preserving and does not affect the theoretical results.} and then proceed to the aggregation step, which relies on the exact quotient semantics of forward
differential equivalence and on the reference-model construction
underlying approximate FDE.

Let $q$ be the polynomial ODE encoding of the trained network and let
$P=\{P_1,\dots,P_k\}$ be the coarsest $\varepsilon$-FDE partition computed on $q$.
By the theory of approximate FDE, there exists a \emph{reference model}
$\hat{q}$, obtained by a bounded perturbation of the coefficients of $q$,
such that the same partition $P$ is an \emph{exact} FDE for $\hat{q}$~\cite{cardelli2023formal}.

For each block $P_m \in P$, define the aggregated (block-sum) variable
\[
y_m = \sum_{x_i \in P_m} x_i.
\]
Since $P$ is an exact FDE for $\hat{q}$, the block-sums satisfy a closed
quotient dynamics
\[
\frac{d y}{d t} = g(y),
\]
where $g$ is the quotient polynomial induced by $\hat{q}$.
We then construct an aggregated neural network $\mathcal{N}_{\text{agg}}$
by replacing each block $P_m$ with a single aggregated neuron representing $y_m$
and choosing its parameters so that the polynomial ODE encoding of
$\mathcal{N}_{\text{agg}}$ coincides with the quotient dynamics $g$.
Our guarantees rely on the fact that $q$ and $\hat{q}$ have close trajectories,
while the quotient of $\hat{q}$ is exact by construction.
\paragraph*{Remark (Approximation with respect to the original model).}
The reference model $\hat{q}$ differs from the original encoding $q$ by a
perturbation of polynomial coefficients whose magnitude is bounded by
the tolerance $\varepsilon$.
As shown in~\cite{cardelli2023formal}, this implies that the trajectories
of $q$ and $\hat{q}$ remain $\mathcal{O}(\varepsilon)$-close.
Consequently, the quotient trajectory $y(t)$ provides an approximation
of the block-sums of the original encoded system with an error
proportional to $\varepsilon$.
Under standard regularity assumptions on the network layers, this
approximation propagates to the input--output behavior, yielding an
$\mathcal{O}(\varepsilon)$ approximation between the outputs of
$\mathcal{N}$ and $\mathcal{N}_{\text{agg}}$.
\paragraph{Remark (Computing the $\varepsilon$-FDE partition)}
Starting from the polynomial ODE encoding in monomial normal form, we compute the coarsest $\varepsilon$-FDE partition refining a given initial partition via \texttt{ERODE}~\cite{cardelli2017erode}, a tool for the analysis and reduction of ODE systems. In this context, ERODE acts as a backend that identifies sets of neurons whose encoded ODE dynamics can be safely aggregated up to a user-specified tolerance $\varepsilon$, which directly governs the degree of aggregation.

\subsection*{Working example}
To make this concrete, consider a feedforward neural network with a single hidden layer and an output.
The network maps two inputs to two hidden neurons with quadratic activation, followed
by a linear output neuron. Let the input be
\(x=(x_1,x_2)^\top \in \mathbb{R}^2\), and define the hidden-layer activations
\[
h_1 = (1.00\,x_1 + 0.50\,x_2 + 0.10)^2, \qquad
h_2 = (1.02\,x_1 + 0.48\,x_2 + 0.12)^2.
\]
The output neuron computes
\[
o = 1.00\,h_1 + 1.00\,h_2 + 0.58 .
\]
Following our encoding, the forward computation is represented as a polynomial ODE:
\begin{align}
\frac{d h_1}{d t} &= (1.00\,x_1 + 0.50\,x_2 + 0.10)^2, \\
\frac{d h_2}{d t} &= (1.02\,x_1 + 0.48\,x_2 + 0.12)^2, \\
\frac{d o}{d t}   &= h_1 + h_2 + 0.58 .
\end{align}

Expanding the right-hand sides and expressing each quadratic term as a
product of variables yields the corresponding coefficients in monomial normal form~\cite{cardelli2023formal,cardelli2016symbolic}.

\begin{align*}
\frac{d h_1}{d t} &= 1.00\,x_1 x_1 + 1.00\,x_1 x_2 + 0.25\,x_2 x_2
                   + 0.20\,x_1 + 0.10\,x_2 + 0.01, \\
\frac{d h_2}{d t} &= 1.04\,x_1 x_1 + 0.98\,x_1 x_2 + 0.23\,x_2 x_2
                   + 0.24\,x_1 + 0.12\,x_2 + 0.01.
\end{align*}

Here, the maximum absolute discrepancy between matching coefficients is
\[
\max \{ |1.04-1.00|,\ |0.98-1.00|,\ |0.23-0.25|,\ |0.24-0.20|,\ |0.12-0.10| \}
= 0.04.
\]
Hence, the block $\{h_1,h_2\}$ is \emph{not} an exact FDE block. Approximate FDE replaces exact equality with a tolerance condition. If the user
selects $\varepsilon < 0.04$, the discrepancy budget is insufficient and the two
neurons remain separate. However, for any choice $\varepsilon \geq 0.04$, the block
$\{h_1,h_2\}$ satisfies the $\varepsilon$-FDE constraints, and the neurons are aggregated.

\paragraph{Effect of aggregation.}
When $\{h_1,h_2\}$ is accepted as an $\varepsilon$-FDE block, the quotient system
tracks the block-sum variable or simply the aggregated neuron $h_{12} = h_1 + h_2$, and the quotient ODE system will be,
\begin{align}
\frac{d h_{12}}{d t} &\approx
(2.04)\,x_1^2 + (1.98)\,x_1 x_2 + (0.48)\,x_2^2
+ (0.44)\,x_1 + (0.22)\,x_2 + 0.02, \\
\frac{d o}{d t} &= h_{12} + 0.58.
\end{align}
By construction, the  polynomial ODE encoding of \(\mathcal{N}_{\text{agg}}\) coincides with this quotient dynamics, and the theory of approximate FDE guarantees that 
$h_{12}$ in the quotient
ODE approximates $h_1+h_2$ of the original system, with an error bounded
linearly in $\varepsilon$~\cite{cardelli2023formal}.

\paragraph{Remark (Interpretation of time).}
The variable $t$ in the ODE encoding does not represent physical or
computational time, nor does it correspond to a forward pass depth.
ODEs are used here as algebraic objects: only the structure of their
right-hand sides matter. References to trajectories and finite time
horizons arise from the theory of approximate differential equivalence
and serve to quantify approximation error, not to model execution
dynamics.

\section{Numerical Evaluation}
\label{sec:numeval}
We consider two classes of datasets: synthetic dynamical-system datasets and real-world public regression datasets.
Our evaluation addresses three questions: (i) how far models can be compressed without loss of predictive accuracy, (ii) whether the resulting compression--accuracy trade-off is stable, and (iii) how the proposed method compares with established pruning baselines under identical conditions.
We choose the tolerance parameter $\varepsilon$ in a scale-aware manner, based on the magnitude of the coefficients in the polynomial ODE encoding. We begin with values of $\varepsilon$ that are small relative to typical coefficient magnitudes, so that only nearly identical dynamics are aggregated, and then gradually increase $\varepsilon$ to permit larger discrepancies consistent with training variability and dataset-specific heterogeneity. In this way, $\varepsilon$ induces a smooth transition from conservative to more aggressive aggregation.
The aggregated models are evaluated on the same test data as the original networks. Predictive accuracy is measured using mean squared error (MSE), enabling us to quantify the trade-off between parameter reduction and output fidelity.
Synthetic datasets are evaluated over 10 random seeds, whereas public datasets are evaluated over 30 random seeds to obtain reliable uncertainty estimates. In all cases, we report 95\% confidence intervals.
\subsection{Synthetic Datasets on Dynamical Systems}

\paragraph{Datasets and setup.}
First, we study the problem of learning derivatives of nonlinear dynamical systems governed by ODEs using neural networks. It provides a known, controlled dynamical experimental environment in which neural networks are trained to learn the derivatives via supervised regression, allowing us to isolate the effects 
of model misspecification.

We evaluate four representative systems selected from prior literature: (i) a reaction network (RN) describing reversible molecular binding~\cite{cardelli2017maximal}, (ii) an inductance-free electrical power distribution network (H-tree)~\cite{rosenfeld2007design,cardelli2023formal}, (iii) a polymerization model (POL) capturing chain-growth dynamics~\cite{cardelli2023formal}, and (iv) a glycolytic oscillator (GLY)~\cite{raissi2018multistep,daniels2015efficient}. With different dimensions, nonlinear structures, and stiffness, these systems encompass biochemical regulation, electrical networks, and chemical kinetics, while providing ground-truth derivatives against which the neural network predictions can be evaluated. For each system, we synthetically generate datasets
$\{(\mathbf{x}_i,\dot{\mathbf{x}}_i)\}_{i=1}^{30000}$ by uniformly sampling states $\mathbf{x}$ over model-specific domains ([0, 100] for all models except the electrical network, and [–25, 25] for the electrical network) and evaluating exact derivatives via the governing equations. The data are split into 70\% training, 20\% validation, and 10\% testing sets, normalized using min–max scaling.

\begin{figure}
    \centering
    \includegraphics[width=\linewidth]{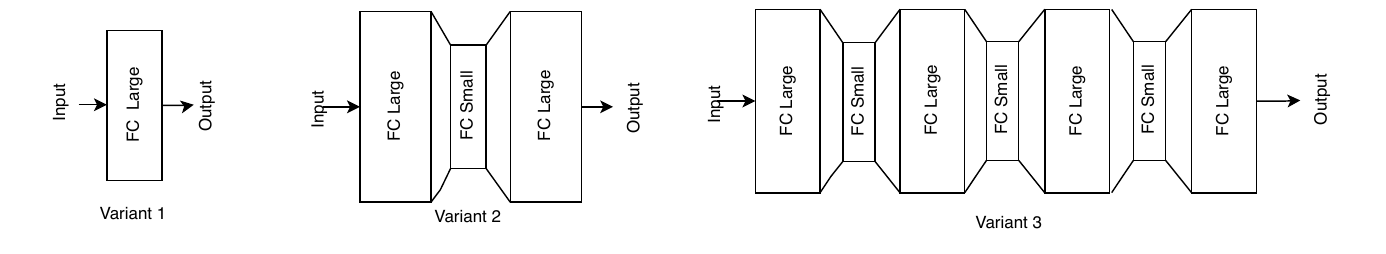}
    \caption{MLP architectures of Variant 1,2,3 models.}
    \label{fig:var123}
\end{figure}

\paragraph{Architectures and training.}
We examine three multilayer perceptron (MLP) variants with increasing depth (Figure~\ref{fig:var123}). 
Variant~1 contains a single large nonlinear FC layer, Variant~2 interleaves two nonlinear FC layers with a linear small FC bottleneck, and Variant~3 extends this pattern to four nonlinear FC layers. The introduction of small FC layers is motivated by scalability
constraints of the \texttt{ERODE} backend: directly encoding large FC
layers leads to polynomial ODE systems that are prohibitively large for
equivalence analysis.
By decomposing wide layers into sequences of smaller ones, we obtain
functionally comparable models whose ODE encodings remain tractable,
allowing aggregation to be applied consistently across variants.
All nonlinear large FC layers employ quadratic activations, enabling expressive polynomial representations while preserving analytical structure.
All models are trained using  Mean Squared Error (MSE) loss function with Adam as optimizer \cite{kingma2014adam} and adaptive learning rate scheduling.
Hyperparameters are selected via grid search (Table~\ref{tab:hyperparams}) and held fixed across pruning methods to ensure fair comparison. 

\begin{table}[t]
\centering
\caption{Hyperparameter grid search ranges.}
\begin{tabular}{ll}
\toprule
\textbf{Hyperparameter} & \textbf{Values} \\ 
\midrule
Epochs & 200, 400 \\
Learning Rate (LR) & 0.1, 0.01, 0.005 \\
Batch Size & 20, 30 \\
Early Stopping Patience & 20, 30, 50 \\
\bottomrule
\end{tabular}
\label{tab:hyperparams}
\end{table}

\paragraph{Evaluation metric and baselines.}\noindent
We use \textbf{Global Remaining Parameters (GRP\%)} as a metric to compute the compression, and define it as the fraction of remaining nonzero parameters after aggregation with respect to the original dense network. Lower GRP\% indicates stronger compression. We compare against two established pruning baselines:
(i) \emph{Magnitude-Based Pruning} (MBP)~\cite{han2015learning}, which removes weights by absolute magnitude,
and (ii) \emph{Wanda}~\cite{sun2023simple}, which includes activation statistics via calibration. For Wanda, we follow the standard 128-sample calibration protocol. 
In our method, the tolerance parameter $\varepsilon$ controls the aggregation, and we set $\varepsilon_1, \varepsilon_3, \varepsilon_5,$ and $\varepsilon_7$ to the values corresponding to the targeted hidden layers for the variants where aggregation is applied. The resulting $\varepsilon$ configurations, together with the pruning
ratios used for MBP and Wanda to achieve comparable compression levels,
are summarized in Table~\ref{tab:epsilon_configs_all_transposed}. For MBP and Wanda, the metric \textbf{Prune\%} is the percentage of pruned weights in the corresponding targeted large hidden layers. All baseline results are averaged over the same random seeds as our method. In Table~\ref{tab:epsilon_configs_all_transposed}, configurations C1–C5 (and C6, where applicable) index progressively more aggressive compression settings, ranging from no compression (C1) to the most aggressive compression levels considered for each dataset and variant.

\begin{table}[!tbp]
\centering
\small
\caption{Our $\varepsilon$ configurations and pruning ratios for (MBP/Wanda) for synthetic datasets across Variants 1--3.}
\label{tab:epsilon_configs_all_transposed}
\renewcommand{\arraystretch}{1.2}
\resizebox{\textwidth}{!}{%
\begin{tabular}{llccccc|cccccc|cccccc|ccccc}
\toprule
& & \multicolumn{5}{c|}{\cellcolor{gray!20}\textbf{RN}} 
& \multicolumn{6}{c|}{\cellcolor{blue!10}\textbf{Electrical Network}} 
& \multicolumn{6}{c|}{\cellcolor{green!15}\textbf{Polymerization}} 
& \multicolumn{5}{c}{\cellcolor{yellow!20}\textbf{Glycolytic}} \\
\cmidrule(lr){3-7}\cmidrule(lr){8-13}\cmidrule(lr){14-19}\cmidrule(lr){20-24}
\textbf{Variant} & \textbf{Method}
& \textbf{C1} & \textbf{C2} & \textbf{C3} & \textbf{C4} & \textbf{C5}
& \textbf{C1} & \textbf{C2} & \textbf{C3} & \textbf{C4} & \textbf{C5} & \textbf{C6}
& \textbf{C1} & \textbf{C2} & \textbf{C3} & \textbf{C4} & \textbf{C5} & \textbf{C6}
& \textbf{C1} & \textbf{C2} & \textbf{C3} & \textbf{C4} & \textbf{C5} \\
\midrule

\multirow{3}{*}{1}
& \makecell[l]{\textbf{Our}($\varepsilon_1$)}
& 0.0 & 0.1 & 0.2 & 0.5 & 1.0
& 0.0 & 0.1 & 0.2 & 0.3 & 0.5 & 1.0
& 0.0 & 0.2 & 0.3 & 0.5 & 1.0 & 2.0
& 0.0 & 0.1 & 0.2 & 0.5 & 1.5 \\
& MBP (Prune\%)
& 0 & 70 & 80 & 85 & 99
& 0 & 60 & 80 & 92 & 97 & 98
& 0 & 60 & 80 & 85 & 95 & 98
& 0 & 45 & 85 & 90 & 99 \\
& Wanda (Prune\%)
& 0 & 70 & 80 & 85 & 99
& 0 & 60 & 80 & 90 & 95 & 99
& 0 & 60 & 80 & 85 & 95 & 99
& 0 & 45 & 85 & 90 & 99 \\
\midrule

\multirow{4}{*}{2}
& \multirow{2}{*}{\makecell[l]{\textbf{Our}($\varepsilon_1$\\    \hspace{20pt}$\varepsilon_3$)}}
& 0.0 & 0.2 & 0.2 & 0.5 & 1.0
& 0.0 & 0.1 & 0.1 & 0.2 & 0.2 & 0.5
& 0.0 & 0.3 & 0.3 & 1.0 & 1.0 & 2.5
& 0.0 & 0.2 & 0.5 & 0.5 & 2.0 \\
& 
& 0.0 & 0.2 & 0.5 & 0.5 & 1.0
& 0.0 & 0.1 & 0.2 & 0.2 & 0.5 & 0.5
& 0.0 & 0.3 & 1.0 & 0.3 & 1.0 & 1.0
& 0.0 & 0.2 & 0.2 & 0.5 & 1.5 \\
& MBP (Prune\%)
& 0 & 50 & 70 & 85 & 99
& 0 & 80 & 85 & 90 & 95 & 99
& 0 & 45 & 80 & 85 & 95 & 99
& 0 & 60 & 85 & 95 & 98 \\
& Wanda (Prune\%)
& 0 & 50 & 70 & 85 & 99
& 0 & 80 & 85 & 90 & 95 & 99
& 0 & 45 & 80 & 85 & 95 & 99
& 0 & 60 & 85 & 95 & 99 \\
\midrule

\multirow{6}{*}{3}
& \multirow{4}{*}{\makecell[l]{\textbf{Our}($\varepsilon_1$\\\hspace{20pt}$\varepsilon_3$\\\hspace{20pt}$\varepsilon_5$\\\hspace{20pt}$\varepsilon_7$)}}
& 0.0 & 0.2 & 0.2 & 0.5 & 1.0
& 0.0 & 0.1 & 0.1 & 0.1 & 0.5 & 0.5
& 0.0 & 0.5 & 0.5 & 0.5 & 1.0 & 1.5
& 0.0 & 0.5 & 0.5 & 0.5 & 0.5 \\
& 
& 0.0 & 0.2 & 0.2 & 0.5 & 1.5
& 0.0 & 0.1 & 0.1 & 0.1 & 0.5 & 0.5
& 0.0 & 0.5 & 0.5 & 0.5 & 1.5 & 2.0
& 0.0 & 0.5 & 0.5 & 0.5 & 0.5 \\
& 
& 0.0 & 0.5 & 0.5 & 0.5 & 1.0
& 0.0 & 0.1 & 0.1 & 0.5 & 0.5 & 0.5
& 0.0 & 0.5 & 0.5 & 1.0 & 1.0 & 2.5
& 0.0 & 0.5 & 0.5 & 1.0 & 1.5 \\
& 
& 0.0 & 0.2 & 0.5 & 0.5 & 1.0
& 0.0 & 0.1 & 0.5 & 0.5 & 0.5 & 1.5
& 0.0 & 0.5 & 1.0 & 1.0 & 1.5 & 2.5
& 0.0 & 0.5 & 1.0 & 1.5 & 2.0 \\
& MBP (Prune\%)
& 0 & 70 & 92 & 97 & 99
& 0 & 80 & 85 & 90 & 95 & 99
& 0 & 80 & 90 & 95 & 97 & 99
& 0 & 50 & 54 & 97 & 99 \\
& Wanda (Prune\%)
& 0 & 70 & 90 & 95 & 99
& 0 & 80 & 85 & 90 & 95 & 99
& 0 & 70 & 80 & 90 & 95 & 99
& 0 & 50 & 70 & 85 & 90 \\
\bottomrule
\end{tabular}
}
\end{table}

\paragraph{Pruning evaluation and comparative analysis.}
Results show a consistent pattern across all synthetic systems and architectural variants (Tables~\ref{tab:rn_all_variants_comparison}--\ref{tab:glycolytic_all_variants_comparison}). The proposed approach maintains very low MSE under aggressive compression
(e.g., beyond 50\% parameter removal), typically outperforming MBP and
Wanda by orders of magnitude at comparable GRP\% across the evaluated
systems and architectural variants. Notably, in the most aggressive configurations (e.g., C5 or C6 as applicable), MBP and Wanda are unable to reduce parameters as our method does, despite being configured with maximal compression targets (e.g., up to 99\%; see Table~3). 
Even in this extreme compression regime, our method achieves orders of magnitude lower MSEs than both baselines.
 Overall, the results show that our method achieves superior resilience and compression performance, with substantially lower MSE at significantly lower GRP.
Importantly, this advantage persists across network depth. Depth often enhances the regime in which significant compression is stable, rather than decreasing the efficiency of aggregation. This suggests that the method leverages functional redundancy rather than architectural characteristics.

\begin{table}[t]
\centering
\small
\caption{Pruning Results Comparison for RN Dataset for all Variants, highlighting lowest GRP\% and mean MSE.}
\label{tab:rn_all_variants_comparison}
\resizebox{\textwidth}{!}{%
\begin{tabular}{c@{\hskip 1em}c@{\hskip 1em}cc@{\hskip 1em}cc@{\hskip 1em}cc}
\toprule
\textbf{Variant} & \textbf{Config} 
& \multicolumn{2}{c}{\textbf{Our Method}} 
& \multicolumn{2}{c}{\textbf{MBP}} 
& \multicolumn{2}{c}{\textbf{Wanda}} \\
\cmidrule(lr){3-4}\cmidrule(lr){5-6}\cmidrule(lr){7-8}
 &  
 & \textbf{GRP\%} & \textbf{MSE $\pm$ 95\% CI} 
 & \textbf{GRP\%} & \textbf{MSE $\pm$ 95\% CI} 
 & \textbf{GRP\%} & \textbf{MSE $\pm$ 95\% CI} \\
\midrule

\multirow{6}{*}{1}
 & C1 & \textbf{\textcolor{blue}{100.00}} & \textbf{\textcolor{red}{6.78e-11 $\pm$ 1.47e-10}} & 100.00 & 6.78e-11 $\pm$ 1.47e-10 & 100.00 & 6.78e-11 $\pm$ 1.47e-10 \\
 & C2 & 67.88 & \textbf{\textcolor{red}{2.64e-09 $\pm$ 1.23e-09}} & 67.82 & 1.25e-04 $\pm$ 9.25e-05 & \textbf{\textcolor{blue}{67.82}} & 3.52e-05 $\pm$ 3.90e-05 \\
 & C3 & \textbf{\textcolor{blue}{51.99}} & \textbf{\textcolor{red}{1.09e-07 $\pm$ 6.33e-08}} & 63.22 & 6.85e-04 $\pm$ 4.12e-04 & 63.22 & 3.13e-04 $\pm$ 2.43e-04 \\
 & C4 & \textbf{\textcolor{blue}{35.52}} & \textbf{\textcolor{red}{1.62e-05 $\pm$ 1.15e-05}} & 60.92 & 2.10e-03 $\pm$ 1.02e-03 & 60.92 & 9.63e-04 $\pm$ 6.06e-04 \\
 & C5 & \textbf{\textcolor{blue}{17.21}} & \textbf{\textcolor{red}{4.12e-03 $\pm$ 5.27e-03}} & 54.54 & 8.39e-02 $\pm$ 1.10e-02 & 54.78 & 5.22e-02 $\pm$ 1.27e-02 \\
\midrule

\multirow{6}{*}{2}
 & C1 & \textbf{\textcolor{blue}{100.00}} & \textbf{\textcolor{red}{1.49e-08 $\pm$ 2.73e-08}} & 100.00 & 1.49e-08 $\pm$ 2.73e-08 & 100.00 & 1.49e-08 $\pm$ 2.73e-08 \\
 & C2 & 79.36 & \textbf{\textcolor{red}{1.65e-08 $\pm$ 2.73e-08}} & 77.01 & 3.76e-05 $\pm$ 3.38e-05 & \textbf{\textcolor{blue}{77.01}} & 2.70e-08 $\pm$ 3.44e-08 \\
 & C3 & \textbf{\textcolor{blue}{59.23}} & \textbf{\textcolor{red}{2.90e-07 $\pm$ 4.18e-07}} & 67.82 & 4.14e-04 $\pm$ 4.41e-04 & 67.82 & 1.51e-05 $\pm$ 2.84e-05 \\
 & C4 & \textbf{\textcolor{blue}{32.34}} & \textbf{\textcolor{red}{1.70e-06 $\pm$ 1.63e-06}} & 60.92 & 3.40e-03 $\pm$ 2.76e-03 & 60.92 & 1.03e-03 $\pm$ 1.60e-03 \\
 & C5 & \textbf{\textcolor{blue}{17.22}} & \textbf{\textcolor{red}{5.85e-04 $\pm$ 6.01e-04}} & 54.53 & 8.04e-02 $\pm$ 2.93e-02 & 54.53 & 5.76e-02 $\pm$ 1.91e-02 \\
\midrule

\multirow{5}{*}{3}
 & C1 & \textbf{\textcolor{blue}{100.00}} & \textbf{\textcolor{red}{2.05e-08 $\pm$ 2.58e-09}} & 100.00 & 2.05e-08 $\pm$ 2.58e-09 & 100.00 & 2.05e-08 $\pm$ 2.58e-09 \\
 & C2 & 67.84 & \textbf{\textcolor{red}{5.92e-08 $\pm$ 6.35e-08}} & 67.82 & 4.14e-04 $\pm$ 1.89e-04 & \textbf{\textcolor{blue}{67.82}} & 1.99e-07 $\pm$ 1.90e-07 \\
 & C3 & \textbf{\textcolor{blue}{57.00}} & \textbf{\textcolor{red}{1.62e-07 $\pm$ 1.00e-07}} & 57.73 & 1.18e-02 $\pm$ 3.43e-03 & 57.73 & 1.92e-03 $\pm$ 1.43e-03 \\
 & C4 & \textbf{\textcolor{blue}{29.65}} & \textbf{\textcolor{red}{3.95e-07 $\pm$ 1.68e-07}} & 55.43 & 4.90e-02 $\pm$ 1.09e-02 & 56.32 & 1.70e-02 $\pm$ 6.02e-03 \\
 & C5 & \textbf{\textcolor{blue}{14.44}} & \textbf{\textcolor{red}{9.78e-04 $\pm$ 9.19e-04}} & 54.43 & 8.72e-02 $\pm$ 2.22e-02 & 54.43 & 5.98e-02 $\pm$ 1.43e-02 \\
\bottomrule
\end{tabular}}
\end{table}

\begin{table}[t]
\centering
\small
\caption{Pruning Results Comparison for Electrical Network Dataset for all Variants, highlighting lowest GRP\% and mean MSE.}
\label{tab:en_all_variants_comparison}
\resizebox{\textwidth}{!}{%
\begin{tabular}{c@{\hskip 1em}c@{\hskip 1em}cc@{\hskip 1em}cc@{\hskip 1em}cc}
\toprule
\textbf{Variant} & \textbf{Config} 
& \multicolumn{2}{c}{\textbf{Our Method}} 
& \multicolumn{2}{c}{\textbf{MBP}} 
& \multicolumn{2}{c}{\textbf{Wanda}} \\
\cmidrule(lr){3-4}\cmidrule(lr){5-6}\cmidrule(lr){7-8}
 &  
 & \textbf{GRP\%} & \textbf{MSE $\pm$ 95\% CI} 
 & \textbf{GRP\%} & \textbf{MSE $\pm$ 95\% CI} 
 & \textbf{GRP\%} & \textbf{MSE $\pm$ 95\% CI} \\
\midrule

\multirow{6}{*}{1}
 & C1 & \textbf{\textcolor{blue}{100.00}} & \textbf{\textcolor{red}{4.89e-11 $\pm$ 1.21e-10}} & 100.00 & 4.89e-11 $\pm$ 1.21e-10 & 100.00 & 4.89e-11 $\pm$ 1.21e-10 \\
 & C2 & \textbf{\textcolor{blue}{71.37}} & \textbf{\textcolor{red}{3.58e-09 $\pm$ 1.44e-09}} & 72.10 & 7.82e-05 $\pm$ 5.33e-05 & 72.10 & 1.54e-05 $\pm$ 1.21e-05 \\
 & C3 & \textbf{\textcolor{blue}{63.88}} & \textbf{\textcolor{red}{4.20e-08 $\pm$ 2.02e-08}} & 62.81 & 1.73e-03 $\pm$ 9.88e-04 & 62.81 & 6.54e-04 $\pm$ 4.92e-04 \\
 & C4 & 57.64 & \textbf{\textcolor{red}{4.44e-07 $\pm$ 2.61e-07}} & \textbf{\textcolor{blue}{57.28}} & 9.03e-03 $\pm$ 4.77e-03 & 58.16 & 4.70e-03 $\pm$ 3.41e-03 \\
 & C5 & \textbf{\textcolor{blue}{47.82}} & \textbf{\textcolor{red}{6.83e-06 $\pm$ 3.74e-06}} & 54.95 & 3.05e-02 $\pm$ 1.08e-02 & 55.84 & 1.37e-02 $\pm$ 6.32e-03 \\
 & C6 & \textbf{\textcolor{blue}{33.26}} & \textbf{\textcolor{red}{2.14e-04 $\pm$ 1.09e-04}} & 54.45 & 4.42e-02 $\pm$ 1.77e-02 & 54.29 & 3.75e-02 $\pm$ 1.52e-02 \\
\midrule

\multirow{6}{*}{2}
 & C1 & \textbf{\textcolor{blue}{100.00}} & \textbf{\textcolor{red}{1.15e-09 $\pm$ 2.01e-09}} & 100.00 & 1.15e-09 $\pm$ 2.01e-09 & 100.00 & 1.15e-09 $\pm$ 2.01e-09 \\
 & C2 & \textbf{\textcolor{blue}{62.05}} & \textbf{\textcolor{red}{1.57e-09 $\pm$ 2.13e-09}} & 62.68 & 2.19e-03 $\pm$ 1.36e-03 & 62.81 & 6.59e-04 $\pm$ 5.02e-04 \\
 & C3 & \textbf{\textcolor{blue}{58.26}} & \textbf{\textcolor{red}{2.38e-09 $\pm$ 2.47e-09}} & 60.36 & 4.04e-03 $\pm$ 2.21e-03 & 60.49 & 2.09e-03 $\pm$ 1.42e-03 \\
 & C4 & \textbf{\textcolor{blue}{53.23}} & \textbf{\textcolor{red}{3.37e-09 $\pm$ 3.11e-09}} & 58.03 & 1.11e-02 $\pm$ 5.42e-03 & 58.16 & 6.69e-03 $\pm$ 3.77e-03 \\
 & C5 & \textbf{\textcolor{blue}{46.70}} & \textbf{\textcolor{red}{5.56e-08 $\pm$ 4.62e-08}} & 57.15 & 1.52e-02 $\pm$ 6.88e-03 & 55.84 & 2.16e-02 $\pm$ 9.73e-03 \\
 & C6 & \textbf{\textcolor{blue}{39.79}} & \textbf{\textcolor{red}{2.33e-07 $\pm$ 1.95e-07}} & 53.88 & 6.83e-02 $\pm$ 2.21e-02 & 54.29 & 5.13e-02 $\pm$ 1.97e-02 \\
\midrule

\multirow{6}{*}{3}
 & C1 & \textbf{\textcolor{blue}{100.00}} & \textbf{\textcolor{red}{1.78e-08 $\pm$ 2.12e-09}} & 100.00 & 1.78e-08 $\pm$ 2.12e-09 & 100.00 & 1.78e-08 $\pm$ 2.12e-09 \\
 & C2 & \textbf{\textcolor{blue}{53.60}} & \textbf{\textcolor{red}{1.79e-08 $\pm$ 2.25e-08}} & 62.28 & 6.96e-04 $\pm$ 3.18e-04 & 62.81 & 8.87e-05 $\pm$ 7.41e-05 \\
 & C3 & \textbf{\textcolor{blue}{49.30}} & \textbf{\textcolor{red}{2.07e-08 $\pm$ 2.64e-08}} & 59.96 & 2.75e-03 $\pm$ 1.02e-03 & 60.48 & 7.48e-04 $\pm$ 3.92e-04 \\
 & C4 & \textbf{\textcolor{blue}{41.81}} & \textbf{\textcolor{red}{3.87e-08 $\pm$ 3.11e-08}} & 57.64 & 9.24e-03 $\pm$ 4.26e-03 & 58.16 & 4.92e-03 $\pm$ 2.74e-03 \\
 & C5 & \textbf{\textcolor{blue}{28.22}} & \textbf{\textcolor{red}{7.52e-08 $\pm$ 5.63e-08}} & 55.31 & 2.86e-02 $\pm$ 1.21e-02 & 55.84 & 2.10e-02 $\pm$ 8.64e-03 \\
 & C6 & \textbf{\textcolor{blue}{26.70}} & \textbf{\textcolor{red}{1.49e-06 $\pm$ 1.08e-06}} & 53.48 & 5.21e-02 $\pm$ 1.88e-02 & 54.29 & 4.81e-02 $\pm$ 1.73e-02 \\
\bottomrule
\end{tabular}}
\end{table}

\begin{table}[t]
\centering
\small
\caption{Pruning Results Comparison for Polymerization Dataset for all Variants, highlighting lowest GRP\% and mean MSE.}
\label{tab:polymerization_all_variants_comparison}
\resizebox{\textwidth}{!}{%
\begin{tabular}{c@{\hskip 1em}c@{\hskip 1em}cc@{\hskip 1em}cc@{\hskip 1em}cc}
\toprule
\textbf{Variant} & \textbf{Config} 
& \multicolumn{2}{c}{\textbf{Our Method}} 
& \multicolumn{2}{c}{\textbf{MBP}} 
& \multicolumn{2}{c}{\textbf{Wanda}} \\
\cmidrule(lr){3-4}\cmidrule(lr){5-6}\cmidrule(lr){7-8}
 &  
 & \textbf{GRP\%} & \textbf{MSE $\pm$ 95\% CI} 
 & \textbf{GRP\%} & \textbf{MSE $\pm$ 95\% CI} 
 & \textbf{GRP\%} & \textbf{MSE $\pm$ 95\% CI} \\
\midrule

\multirow{6}{*}{1}
 & C1 & \textbf{\textcolor{blue}{100.00}} & \textbf{\textcolor{red}{7.90e-11 $\pm$ 1.62e-10}} & 100.00 & 7.90e-11 $\pm$ 1.62e-10 & 100.00 & 7.90e-11 $\pm$ 1.62e-10 \\
 & C2 & 73.20 & \textbf{\textcolor{red}{3.79e-09 $\pm$ 1.48e-09}} & 71.69 & 7.55e-05 $\pm$ 5.41e-05 & \textbf{\textcolor{blue}{71.69}} & 1.26e-05 $\pm$ 1.03e-05 \\
 & C3 & 63.46 & \textbf{\textcolor{red}{3.82e-08 $\pm$ 1.96e-08}} & 62.25 & 7.64e-04 $\pm$ 4.51e-04 & \textbf{\textcolor{blue}{62.25}} & 4.19e-04 $\pm$ 2.87e-04 \\
 & C4 & \textbf{\textcolor{blue}{52.89}} & \textbf{\textcolor{red}{3.08e-07 $\pm$ 2.11e-07}} & 59.89 & 1.55e-03 $\pm$ 8.84e-04 & 59.89 & 8.97e-04 $\pm$ 5.71e-04 \\
 & C5 & \textbf{\textcolor{blue}{36.75}} & \textbf{\textcolor{red}{3.63e-05 $\pm$ 2.71e-05}} & 55.17 & 1.25e-02 $\pm$ 6.52e-03 & 55.17 & 6.54e-03 $\pm$ 3.88e-03 \\
 & C6 & \textbf{\textcolor{blue}{16.44}} & \textbf{\textcolor{red}{3.00e-03 $\pm$ 2.19e-03}} & 53.77 & 2.69e-02 $\pm$ 1.14e-02 & 53.60 & 1.88e-02 $\pm$ 8.72e-03 \\
\midrule

\multirow{6}{*}{2}
 & C1 & \textbf{\textcolor{blue}{100.00}} & \textbf{\textcolor{red}{2.55e-09 $\pm$ 3.84e-09}} & 100.00 & 2.55e-09 $\pm$ 3.84e-09 & 100.00 & 2.55e-09 $\pm$ 3.84e-09 \\
 & C2 & 82.98 & \textbf{\textcolor{red}{2.73e-09 $\pm$ 3.91e-09}} & 78.77 & 4.16e-05 $\pm$ 3.61e-05 & \textbf{\textcolor{blue}{78.77}} & 2.69e-08 $\pm$ 3.17e-08 \\
 & C3 & 62.75 & \textbf{\textcolor{red}{2.65e-07 $\pm$ 2.98e-07}} & 62.25 & 1.61e-03 $\pm$ 1.02e-03 & \textbf{\textcolor{blue}{62.25}} & 3.29e-04 $\pm$ 2.44e-04 \\
 & C4 & \textbf{\textcolor{blue}{58.05}} & \textbf{\textcolor{red}{1.22e-05 $\pm$ 1.03e-05}} & 59.89 & 3.48e-03 $\pm$ 1.81e-03 & 59.89 & 1.14e-03 $\pm$ 6.87e-04 \\
 & C5 & \textbf{\textcolor{blue}{37.86}} & \textbf{\textcolor{red}{1.26e-05 $\pm$ 1.11e-05}} & 55.17 & 1.95e-02 $\pm$ 8.42e-03 & 55.17 & 1.31e-02 $\pm$ 6.21e-03 \\
 & C6 & \textbf{\textcolor{blue}{25.57}} & \textbf{\textcolor{red}{5.92e-03 $\pm$ 4.88e-03}} & 53.30 & 4.70e-02 $\pm$ 1.65e-02 & 53.60 & 3.35e-02 $\pm$ 1.27e-02 \\
\midrule

\multirow{6}{*}{3}
 & C1 & \textbf{\textcolor{blue}{100.00}} & \textbf{\textcolor{red}{1.40e-07 $\pm$ 1.72e-07}} & 100.00 & 1.40e-07 $\pm$ 1.72e-07 & 100.00 & 1.40e-07 $\pm$ 1.72e-07 \\
 & C2 & \textbf{\textcolor{blue}{61.19}} & \textbf{\textcolor{red}{1.81e-07 $\pm$ 2.03e-07}} & 62.25 & 1.36e-03 $\pm$ 6.82e-04 & 66.97 & 7.65e-06 $\pm$ 6.94e-06 \\
 & C3 & \textbf{\textcolor{blue}{56.03}} & \textbf{\textcolor{red}{3.99e-07 $\pm$ 3.11e-07}} & 57.54 & 6.87e-03 $\pm$ 2.41e-03 & 62.25 & 8.39e-05 $\pm$ 5.92e-05 \\
 & C4 & \textbf{\textcolor{blue}{45.25}} & \textbf{\textcolor{red}{4.12e-07 $\pm$ 3.44e-07}} & 55.18 & 1.72e-02 $\pm$ 6.54e-03 & 57.54 & 1.99e-03 $\pm$ 1.02e-03 \\
 & C5 & \textbf{\textcolor{blue}{25.17}} & \textbf{\textcolor{red}{7.37e-06 $\pm$ 6.02e-06}} & 54.26 & 2.47e-02 $\pm$ 9.11e-03 & 55.18 & 1.10e-02 $\pm$ 5.08e-03 \\
 & C6 & \textbf{\textcolor{blue}{17.37}} & \textbf{\textcolor{red}{4.32e-04 $\pm$ 3.66e-04}} & 53.29 & 3.46e-02 $\pm$ 1.22e-02 & 53.60 & 3.18e-02 $\pm$ 1.14e-02 \\
\bottomrule
\end{tabular}}
\end{table}

\begin{table}[t]
\centering
\small
\caption{Pruning Results Comparison for Glycolytic Dataset for all Variants, highlighting lowest GRP\% and mean MSE.}
\label{tab:glycolytic_all_variants_comparison}
\resizebox{\textwidth}{!}{%
\begin{tabular}{c@{\hskip 1em}c@{\hskip 1em}cc@{\hskip 1em}cc@{\hskip 1em}cc}
\toprule
\textbf{Variant} & \textbf{Config} 
& \multicolumn{2}{c}{\textbf{Our Method}} 
& \multicolumn{2}{c}{\textbf{MBP}} 
& \multicolumn{2}{c}{\textbf{Wanda}} \\
\cmidrule(lr){3-4}\cmidrule(lr){5-6}\cmidrule(lr){7-8}
 &  
 & \textbf{GRP\%} & \textbf{MSE $\pm$ 95\% CI} 
 & \textbf{GRP\%} & \textbf{MSE $\pm$ 95\% CI} 
 & \textbf{GRP\%} & \textbf{MSE $\pm$ 95\% CI} \\
\midrule

\multirow{5}{*}{1}
 & C1 & \textbf{\textcolor{blue}{100.00}} & \textbf{\textcolor{red}{2.03e-04 $\pm$ 1.91e-05}} & 100.00 & 2.03e-04 $\pm$ 1.91e-05 & 100.00 & 2.03e-04 $\pm$ 1.91e-05 \\
 & C2 & \textbf{\textcolor{blue}{75.28}} & \textbf{\textcolor{red}{2.03e-04 $\pm$ 1.88e-05}} & 79.08 & 2.04e-04 $\pm$ 2.02e-05 & 79.08 & 2.04e-04 $\pm$ 2.02e-05 \\
 & C3 & \textbf{\textcolor{blue}{47.99}} & \textbf{\textcolor{red}{2.04e-04 $\pm$ 1.95e-05}} & 60.49 & 8.69e-04 $\pm$ 3.41e-04 & 60.49 & 3.73e-04 $\pm$ 2.11e-04 \\
 & C4 & \textbf{\textcolor{blue}{29.51}} & \textbf{\textcolor{red}{2.04e-04 $\pm$ 1.99e-05}} & 58.16 & 2.43e-03 $\pm$ 8.74e-04 & 58.16 & 1.03e-03 $\pm$ 4.98e-04 \\
 & C5 & \textbf{\textcolor{blue}{17.78}} & \textbf{\textcolor{red}{4.95e-04 $\pm$ 2.72e-04}} & 54.01 & 8.62e-02 $\pm$ 2.11e-02 & 54.28 & 2.54e-02 $\pm$ 1.12e-02 \\
\midrule

\multirow{5}{*}{2}
 & C1 & \textbf{\textcolor{blue}{100.00}} & \textbf{\textcolor{red}{1.75e-04 $\pm$ 1.66e-05}} & 100.00 & 1.75e-04 $\pm$ 1.66e-05 & 100.00 & 1.75e-04 $\pm$ 1.66e-05 \\
 & C2 & 75.62 & \textbf{\textcolor{red}{1.75e-04 $\pm$ 1.64e-05}} & 72.11 & 4.50e-04 $\pm$ 2.28e-04 & \textbf{\textcolor{blue}{72.11}} & 1.80e-04 $\pm$ 1.73e-05 \\
 & C3 & \textbf{\textcolor{blue}{56.43}} & \textbf{\textcolor{red}{1.76e-04 $\pm$ 1.68e-05}} & 60.49 & 3.93e-03 $\pm$ 1.21e-03 & 60.49 & 8.74e-04 $\pm$ 3.65e-04 \\
 & C4 & \textbf{\textcolor{blue}{39.94}} & \textbf{\textcolor{red}{1.77e-04 $\pm$ 1.70e-05}} & 55.83 & 2.36e-02 $\pm$ 6.51e-03 & 55.83 & 1.21e-02 $\pm$ 4.42e-03 \\
 & C5 & \textbf{\textcolor{blue}{13.44}} & \textbf{\textcolor{red}{7.79e-03 $\pm$ 3.82e-03}} & 54.45 & 4.09e-02 $\pm$ 1.44e-02 & 54.29 & 4.02e-02 $\pm$ 1.39e-02 \\
\midrule

\multirow{5}{*}{3}
 & C1 & \textbf{\textcolor{blue}{100.00}} & \textbf{\textcolor{red}{6.32e-04 $\pm$ 4.88e-05}} & 100.00 & 6.30e-04 $\pm$ 4.81e-05 & 100.00 & 6.30e-04 $\pm$ 4.81e-05 \\
 & C2 & \textbf{\textcolor{blue}{76.66}} & \textbf{\textcolor{red}{6.35e-04 $\pm$ 5.02e-05}} & 76.76 & 6.53e-03 $\pm$ 2.11e-03 & 76.76 & 1.42e-03 $\pm$ 6.72e-04 \\
 & C3 & \textbf{\textcolor{blue}{69.94}} & \textbf{\textcolor{red}{7.60e-04 $\pm$ 6.11e-05}} & 74.88 & 1.42e-02 $\pm$ 4.87e-03 & 67.46 & 5.90e-03 $\pm$ 2.31e-03 \\
 & C4 & \textbf{\textcolor{blue}{55.84}} & \textbf{\textcolor{red}{1.26e-03 $\pm$ 8.22e-05}} & 54.95 & 8.57e-02 $\pm$ 2.41e-02 & 60.49 & 4.02e-02 $\pm$ 1.52e-02 \\
 & C5 & \textbf{\textcolor{blue}{49.63}} & \textbf{\textcolor{red}{9.38e-03 $\pm$ 4.11e-03}} & 54.01 & 9.05e-02 $\pm$ 2.77e-02 & 58.16 & 4.11e-01 $\pm$ 1.32e-01 \\
\bottomrule
\end{tabular}}
\end{table}

\begin{figure}[t]
  \centering
\includegraphics[width=0.48\linewidth]{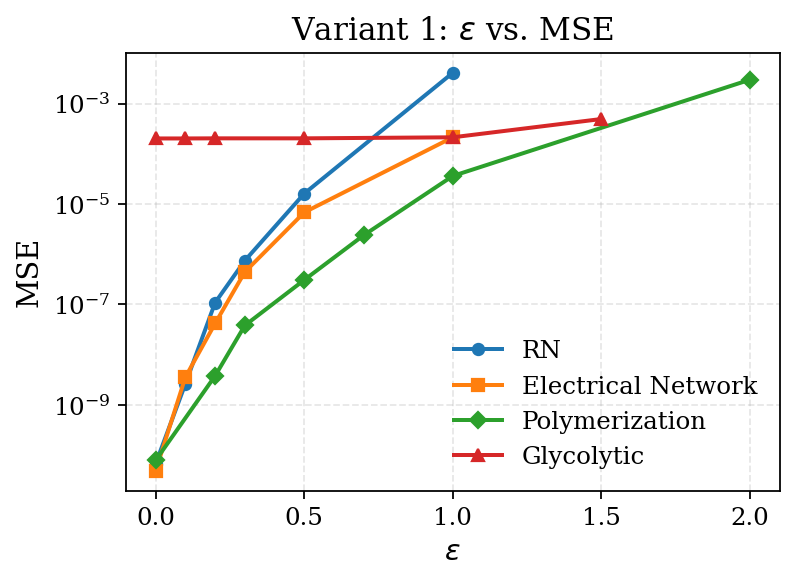}\hfill
\includegraphics[width=0.48\linewidth]{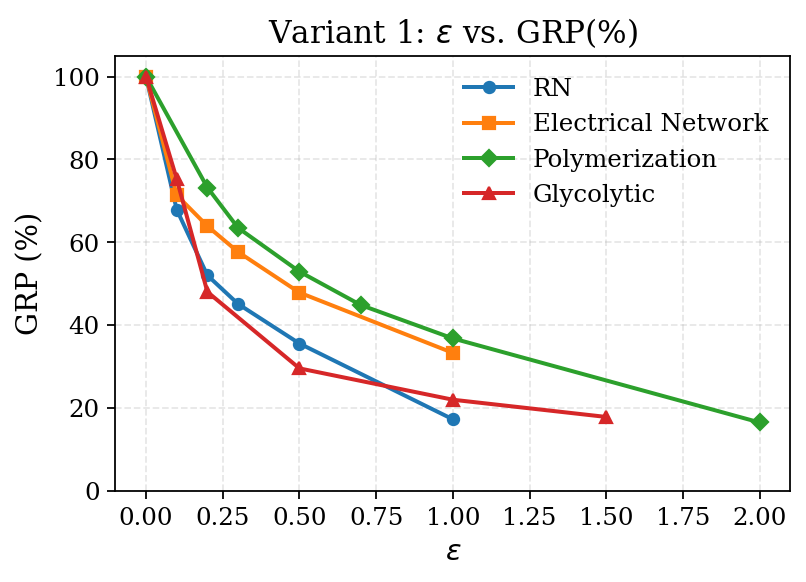}
  \caption{Sensitivity Analysis of $\varepsilon$ across synthetic datasets for variant 1}
  \label{fig:mse-grp-quad}
\end{figure}

\paragraph{Sensitivity to aggregation tolerance.}
Figure~\ref{fig:mse-grp-quad} reports sensitivity curves with respect to the aggregation tolerance~$\varepsilon$, shown for Variant~1 for illustrative
purposes. Increasing~$\varepsilon$ monotonically reduces GRP\% while gradually increasing MSE. For all systems, there exists a broad operating region in which significant compression is achieved with negligible loss in accuracy, followed by a sharper transition where further aggregation becomes detrimental. This behavior demonstrates that~$\varepsilon$ acts as a smooth control parameter rather than a brittle pruning threshold.

\subsection{Public Datasets}

\paragraph{Datasets and protocol.}
We next evaluate the method on four public regression benchmarks: Abalone~\cite{abalone_1}, Metro Interstate Traffic Volume~\cite{metro_interstate_traffic_volume_492}, Individual Household Electric Power Consumption (HEPC)~\cite{individual_household}, and Protein (CASP)~\cite{physicochemical_properties_of_protein_tertiary_structure_265}. These datasets differ substantially in size, feature composition, noise characteristics, and target scale, providing a stringent test of generalization beyond equation-generated data.

We process all datasets using a consistent pipeline. All inputs are converted to numerical representations, and continuous features are standardized using the training-set mean and standard 
deviation; the same transformation is applied to validation and test
data. Targets are standardized using training-set statistics for
optimization and inverse-transformed for reporting. For datasets containing temporal information (Metro and HEPC), we convert timestamp features into numerical features. Across all public-dataset experiments, we use variant 1 to isolate the effect of pruning from architectural depth. The resulting $\varepsilon$ configurations, together with the pruning ratios used for MBP and Wanda to achieve comparable compression levels, are summarized in Table~\ref{tab:epsilon_configs_public_transposed}. 

\begin{table}[t]
\centering
\small
\caption{Our $\varepsilon$ configurations and pruning ratios (MBP/Wanda) for public datasets across Variant~1.}
\label{tab:epsilon_configs_public_transposed}
\renewcommand{\arraystretch}{1.3}
\resizebox{\textwidth}{!}{%
\begin{tabular}{llccccccc|ccccccc|ccccccc|ccccccc}
\toprule
& & \multicolumn{7}{c|}{\cellcolor{gray!20}\textbf{HEPC}} 
& \multicolumn{7}{c|}{\cellcolor{blue!20}\textbf{Abalone}} 
& \multicolumn{7}{c|}{\cellcolor{green!20}\textbf{Metro}} 
& \multicolumn{7}{c}{\cellcolor{yellow!20}\textbf{Protein}} \\
\cmidrule(lr){3-9}\cmidrule(lr){10-16}\cmidrule(lr){17-23}\cmidrule(lr){24-30}
\textbf{Variant} & \textbf{Method}
& \textbf{C1} & \textbf{C2} & \textbf{C3} & \textbf{C4} & \textbf{C5} & \textbf{C6} & \textbf{C7}
& \textbf{C1} & \textbf{C2} & \textbf{C3} & \textbf{C4} & \textbf{C5} & \textbf{C6} & \textbf{C7}
& \textbf{C1} & \textbf{C2} & \textbf{C3} & \textbf{C4} & \textbf{C5} & \textbf{C6} & \textbf{C7}
& \textbf{C1} & \textbf{C2} & \textbf{C3} & \textbf{C4} & \textbf{C5} & \textbf{C6} & \textbf{C7} \\
\midrule

\multirow{3}{*}{1}
& Our ($\varepsilon_1$)
& 0.0 & 0.001 & 0.002 & 0.003 & 0.005 & 0.01 & 0.03
& 0.0 & 0.003 & 0.005 & 0.007 & 0.01 & 0.02 & 0.03
& 0.0 & 0.003 & 0.005 & 0.006 & 0.009 & 0.02 & 0.04
& 0.0 & 0.007 & 0.01 & 0.02 & 0.03 & 0.05 & 0.09 \\

& MBP (Prune\%)
& 0 & 30 & 50 & 70 & 80 & 90 & 98
& 0 & 20.83 & 39.97 & 58.22 & 74.08 & 87.94 & 97.53
& 0 & 21.52 & 47.75 & 67.89 & 78.58 & 87.44 & 91.94
& 0 & 25 & 55.44 & 73.56 & 82.44 & 91.36 & 95.11 \\

& Wanda (Prune\%)
& 0 & 45 & 50 & 70 & 80 & 90 & 98
& 0 & 21 & 40 & 57 & 74 & 88 & 98
& 0 & 22 & 48 & 68 & 79 & 87 & 95
& 0 & 25 & 55 & 73 & 82 & 91 & 99 \\

\bottomrule
\end{tabular}
}
\end{table}

\begin{table}[t]
\centering
\small
\caption{Pruning Results Comparison for Public Datasets for Variant~1, highlighting lowest GRP\% and mean MSE.}
\label{tab:public_all_datasets_variant1}
\resizebox{\textwidth}{!}{%
\begin{tabular}{c@{\hskip 1em}c@{\hskip 1em}cc@{\hskip 1em}cc@{\hskip 1em}cc}
\toprule
\textbf{Dataset} & \textbf{Config} 
& \multicolumn{2}{c}{\textbf{Our Method}} 
& \multicolumn{2}{c}{\textbf{MBP}} 
& \multicolumn{2}{c}{\textbf{Wanda}} \\
\cmidrule(lr){3-4}\cmidrule(lr){5-6}\cmidrule(lr){7-8}
 &  
 & \textbf{GRP\%} & \textbf{MSE $\pm$ 95\% CI} 
 & \textbf{GRP\%} & \textbf{MSE $\pm$ 95\% CI} 
 & \textbf{GRP\%} & \textbf{MSE $\pm$ 95\% CI} \\
\midrule

\multirow{7}{*}{Abalone}
 & C1 & \textbf{\textcolor{blue}{100.00}} & \textbf{\textcolor{red}{7.25e-03 $\pm$ 1.17e-03}}
 & 100.00 & 7.25e-03 $\pm$ 1.17e-03
 & 100.00 & 7.25e-03 $\pm$ 1.17e-03 \\
 & C2 & \textbf{\textcolor{blue}{70.89}} & 9.94e-03 $\pm$ 1.52e-03
 & 97.92 & 7.38e-03 $\pm$ 1.15e-03
 & 95.36 & \textbf{\textcolor{red}{7.27e-03 $\pm$ 1.17e-03}} \\
 & C3 & \textbf{\textcolor{blue}{57.59}} & 9.96e-03 $\pm$ 1.52e-03
 & 93.50 & 9.03e-03 $\pm$ 1.34e-03
 & 93.49 & \textbf{\textcolor{red}{7.62e-03 $\pm$ 1.43e-03}} \\
 & C4 & \textbf{\textcolor{blue}{46.90}} & 9.92e-03 $\pm$ 1.49e-03
 & 67.72 & 1.83e-02 $\pm$ 4.45e-03
 & 66.93 & \textbf{\textcolor{red}{8.32e-03 $\pm$ 1.27e-03}} \\
 & C5 & \textbf{\textcolor{blue}{35.37}} & \textbf{\textcolor{red}{9.89e-03 $\pm$ 1.52e-03}}
 & 67.29 & 4.58e-02 $\pm$ 8.68e-03
 & 66.31 & 1.03e-02 $\pm$ 1.82e-03 \\
 & C6 & \textbf{\textcolor{blue}{18.13}} & \textbf{\textcolor{red}{1.08e-02 $\pm$ 1.78e-03}}
 & 41.03 & 7.91e-02 $\pm$ 1.14e-02
 & 40.58 & 2.16e-02 $\pm$ 6.20e-03 \\
 & C7 & \textbf{\textcolor{blue}{11.46}} & \textbf{\textcolor{red}{2.00e-02 $\pm$ 5.09e-03}}
 & 40.65 & 1.07e-01 $\pm$ 1.40e-02
 & 40.46 & 6.32e-02 $\pm$ 1.24e-02 \\
\midrule

\multirow{7}{*}{HEPC}
 & C1 & \textbf{\textcolor{blue}{100.00}} & \textbf{\textcolor{red}{1.32e-03 $\pm$ 6.00e-06}}
 & 100.00 & 1.32e-03 $\pm$ 6.00e-06
 & 100.00 & 1.32e-03 $\pm$ 6.00e-06 \\
 & C2 & \textbf{\textcolor{blue}{56.06}} & 4.17e-03 $\pm$ 1.01e-03
 & 66.94 & \textbf{\textcolor{red}{2.02e-03 $\pm$ 1.36e-03}}
 & 65.84 & 2.35e-03 $\pm$ 9.94e-04 \\
 & C3 & \textbf{\textcolor{blue}{50.60}} & 4.17e-03 $\pm$ 1.01e-03
 & 65.46 & 3.83e-03 $\pm$ 2.88e-03
 & 65.44 & \textbf{\textcolor{red}{3.14e-03 $\pm$ 1.62e-03}} \\
 & C4 & \textbf{\textcolor{blue}{45.51}} & \textbf{\textcolor{red}{4.16e-03 $\pm$ 1.00e-03}}
 & 63.61 & 4.17e-02 $\pm$ 1.56e-02
 & 63.61 & 3.35e-02 $\pm$ 2.19e-02 \\
 & C5 & \textbf{\textcolor{blue}{38.02}} & \textbf{\textcolor{red}{4.17e-03 $\pm$ 9.96e-04}}
 & 62.61 & 1.46e-01 $\pm$ 5.64e-02
 & 62.61 & 9.04e-02 $\pm$ 2.86e-02 \\
 & C6 & \textbf{\textcolor{blue}{24.62}} & \textbf{\textcolor{red}{4.62e-03 $\pm$ 1.01e-03}}
 & 61.61 & 3.82e-01 $\pm$ 7.25e-02
 & 61.61 & 2.72e-01 $\pm$ 4.00e-02 \\
 & C7 & \textbf{\textcolor{blue}{7.40}} & \textbf{\textcolor{red}{2.12e-01 $\pm$ 7.70e-02}}
 & 60.86 & 8.43e-01 $\pm$ 4.75e-02
 & 60.86 & 7.15e-01 $\pm$ 5.13e-02 \\
\midrule

\multirow{7}{*}{Metro}
 & C1 & \textbf{\textcolor{blue}{100.00}} & \textbf{\textcolor{red}{2.20e-02 $\pm$ 1.51e-04}}
 & 100.00 & 2.20e-02 $\pm$ 1.51e-04
 & 100.00 & 2.20e-02 $\pm$ 1.51e-04 \\
 & C2 & \textbf{\textcolor{blue}{73.25}} & 2.62e-02 $\pm$ 1.69e-03
 & 95.24 & 2.37e-02 $\pm$ 6.42e-04
 & 95.22 & \textbf{\textcolor{red}{2.21e-02 $\pm$ 1.58e-04}} \\
 & C3 & \textbf{\textcolor{blue}{60.09}} & 2.62e-02 $\pm$ 1.69e-03
 & 93.06 & 9.20e-02 $\pm$ 1.21e-02
 & 93.08 & \textbf{\textcolor{red}{2.38e-02 $\pm$ 5.58e-04}} \\
 & C4 & \textbf{\textcolor{blue}{55.12}} & \textbf{\textcolor{red}{2.62e-02 $\pm$ 1.71e-03}}
 & 64.32 & 3.66e-01 $\pm$ 3.23e-02
 & 63.22 & 3.17e-02 $\pm$ 2.25e-03 \\
 & C5 & \textbf{\textcolor{blue}{43.03}} & \textbf{\textcolor{red}{2.63e-02 $\pm$ 1.77e-03}}
 & 63.43 & 5.52e-01 $\pm$ 3.41e-02
 & 62.73 & 4.59e-02 $\pm$ 3.79e-03 \\
 & C6 & \textbf{\textcolor{blue}{21.12}} & \textbf{\textcolor{red}{3.05e-02 $\pm$ 3.45e-03}}
 & 34.67 & 5.16e-01 $\pm$ 3.12e-02
 & 34.06 & 7.86e-02 $\pm$ 9.39e-03 \\
 & C7 & \textbf{\textcolor{blue}{10.57}} & 1.45e-01 $\pm$ 3.58e-02
 & 34.29 & 4.30e-01 $\pm$ 2.83e-02
 & 33.74 & \textbf{\textcolor{red}{1.17e-01 $\pm$ 1.45e-02}} \\
\midrule

\multirow{7}{*}{Protein}
 & C1 & \textbf{\textcolor{blue}{100.00}} & \textbf{\textcolor{red}{5.61e-02 $\pm$ 4.03e-04}}
 & 100.00 & 5.61e-02 $\pm$ 4.03e-04
 & 100.00 & 5.61e-02 $\pm$ 4.03e-04 \\
 & C2 & \textbf{\textcolor{blue}{73.20}} & 5.97e-02 $\pm$ 1.21e-03
 & 95.33 & 6.16e-02 $\pm$ 2.31e-03
 & 95.33 & \textbf{\textcolor{red}{5.65e-02 $\pm$ 4.25e-04}} \\
 & C3 & \textbf{\textcolor{blue}{64.98}} & \textbf{\textcolor{red}{5.97e-02 $\pm$ 1.22e-03}}
 & 71.46 & 1.22e-01 $\pm$ 1.22e-02
 & 71.05 & 6.20e-02 $\pm$ 1.42e-03 \\
 & C4 & \textbf{\textcolor{blue}{44.42}} & \textbf{\textcolor{red}{5.97e-02 $\pm$ 1.25e-03}}
 & 69.81 & 1.65e-01 $\pm$ 1.81e-02
 & 69.75 & 7.74e-02 $\pm$ 6.43e-03 \\
 & C5 & \textbf{\textcolor{blue}{32.51}} & \textbf{\textcolor{red}{5.98e-02 $\pm$ 1.40e-03}}
 & 69.00 & 1.77e-01 $\pm$ 1.69e-02
 & 69.00 & 9.15e-02 $\pm$ 8.47e-03 \\
 & C6 & \textbf{\textcolor{blue}{21.32}} & \textbf{\textcolor{red}{6.09e-02 $\pm$ 2.82e-03}}
 & 45.50 & 1.89e-01 $\pm$ 1.56e-02
 & 45.29 & 1.24e-01 $\pm$ 1.51e-02 \\
 & C7 & \textbf{\textcolor{blue}{12.99}} & \textbf{\textcolor{red}{9.59e-02 $\pm$ 1.27e-02}}
 & 44.98 & 1.87e-01 $\pm$ 1.50e-02
 & 44.83 & 1.79e-01 $\pm$ 1.63e-02 \\
\bottomrule
\end{tabular}}
\end{table}

\paragraph{Pruning evaluation and comparative analysis.}
Table~\ref{tab:public_all_datasets_variant1} summarizes the results. Across all datasets, the proposed method consistently achieves lower error at comparable or lower GRP\% relative to MBP and Wanda. More importantly, the degradation in performance as compression increases is substantially smoother. For example, on HEPC, the proposed method maintains MSE on the order of $10^{-3}$ even when GRP falls below 30\%, whereas baseline methods incur errors that increase by one to two orders of magnitude at similar compression levels. Similar trends are observed on Metro and Protein, despite their higher variance and more complex feature interactions. Confidence intervals further indicate that isolated favorable seeds do not drive these improvements. In most regimes, the proposed method exhibits uncertainty comparable to or smaller than that of the baselines, suggesting stable behavior under random initialization and data splits.

\begin{figure}[t]
\centering
\includegraphics[width=0.48\linewidth]{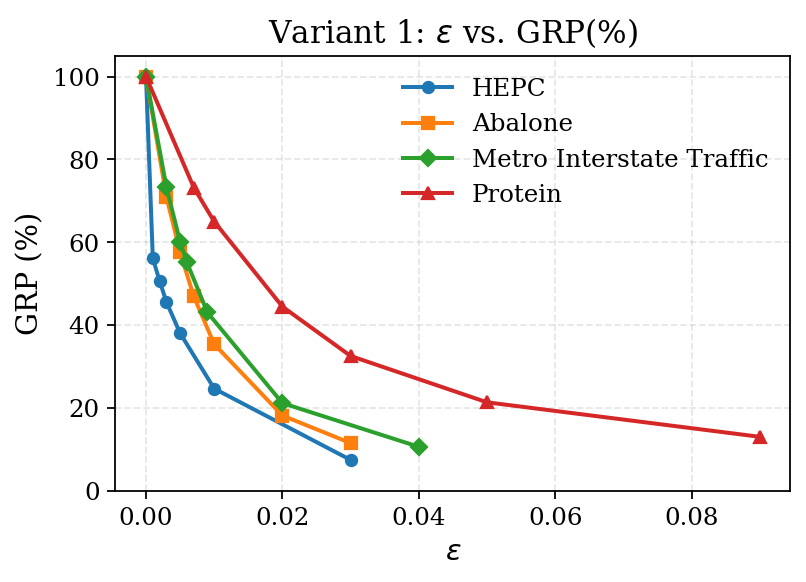}\hfill
\includegraphics[width=0.48\linewidth]{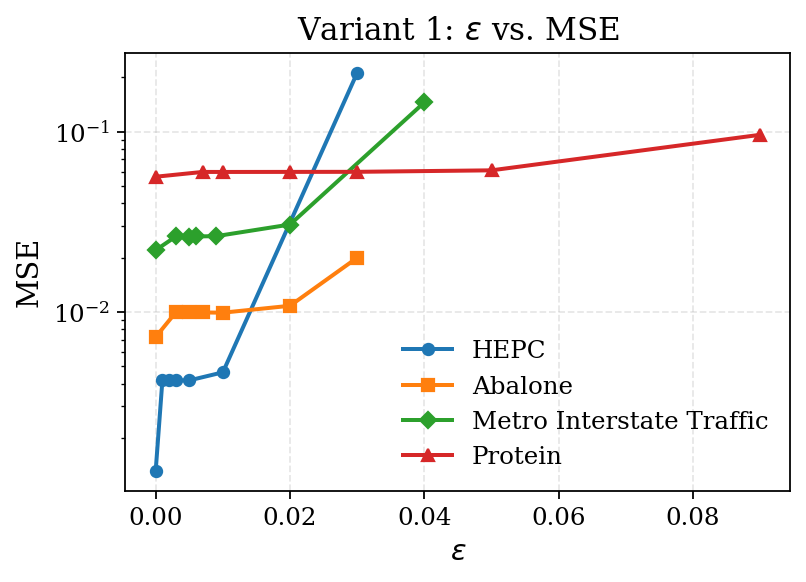}
\caption{Sensitivity Analysis of $\varepsilon$ across public datasets for variant 1}
\label{fig:sens_aggregate_public}
\end{figure}

\paragraph{Sensitivity to aggregation tolerance.}
Figure~\ref{fig:sens_aggregate_public} illustrates the effect of~$\varepsilon$ on GRP\% and MSE across all public benchmarks. As with synthetic systems, each dataset exhibits a dataset-specific compression regime in which substantial parameter reduction is possible with minimal loss in accuracy. Beyond this regime, error increases more rapidly, but remains consistently lower than that of competing methods at matched GRP\%. Aggregate sensitivity views further confirm that the proposed method offers a wider feasible operating region, allowing practitioners to select compression levels based on parameter budgets while retaining predictable performance.

\subsection{Summary of Empirical Findings}
Across both synthetic and real-world datasets, the proposed aggregation-based pruning method demonstrates a consistently superior trade-off between compression and accuracy while preserving learned functional dynamics. Its advantages include stability across architectures, robustness under aggressive compression, and resilience to realistic data noise and heterogeneity. These results suggest that aggregating approximately equivalent functional components provides a principled alternative to traditional pruning strategies that operate solely at the level of individual weights.





\section{Conclusion and Future Work}
This paper studied neural network compression through approximate forward differential equivalence. Instead of removing parameters based on local importance, the proposed method aggregates neurons with similar functional behavior, yielding a principled alternative to conventional pruning based on dynamical equivalence.
The numerical results support this perspective in two settings, on synthetic dynamical systems and on public regression benchmarks. In both settings, aggregation consistently preserves accuracy while substantially reducing model size. Relative to pruning baselines, these results suggest that aggregating functionally redundant neurons can be more effective than removing parameters independently.
The current framework is nonetheless constrained by its assumptions. It relies on polynomial representations and on encoding neural networks as ODE systems, which limits direct applicability to certain architectures. Moreover, although the tolerance parameter $\varepsilon$ offers an interpretable control of the compression--accuracy trade-off, it currently requires manual tuning and may not transfer uniformly across datasets and model sizes.
Future work includes adaptive strategies for selecting $\varepsilon$, extensions to broader classes of activations and architectures, and integration with other compression techniques for evaluation on hardware-constrained platforms.
\bibliographystyle{splncs04}
\bibliography{allreferences}

%




\end{document}